\documentclass[sigconf]{acmart}

\copyrightyear{2022} 
\acmYear{2022} 
\setcopyright{acmcopyright}\acmConference[CIKM '22]{Proceedings of the 31st ACM
International Conference on Information and Knowledge Management}{October
17--21, 2022}{Atlanta, GA, USA}
\acmBooktitle{Proceedings of the 31st ACM International Conference on Information
and Knowledge Management (CIKM '22), October 17--21, 2022, Atlanta, GA, USA}
\acmPrice{15.00}
\acmDOI{10.1145/3511808.3557648}
\acmISBN{978-1-4503-9236-5/22/10}

\AtBeginDocument{%
  \providecommand\BibTeX{{%
    \normalfont B\kern-0.5em{\scshape i\kern-0.25em b}\kern-0.8em\TeX}}}

\begin{document}

\title{Molecular Substructure-Aware Network for Drug-Drug Interaction Prediction}

\author{Xinyu Zhu}
\affiliation{
  \institution{Zhejiang University}
  \city{Hangzhou}
  \country{China}
}
\email{zxy21@zju.edu.cn}

\author{Yongliang Shen}
\affiliation{
  \institution{Zhejiang University}
  \city{Hangzhou}
  \country{China}
}
\email{syl@zju.edu.cn}

\author{Weiming Lu}
\authornote{Corresponding author.}
\affiliation{
  \institution{Zhejiang University}
  \city{Hangzhou}
  \country{China}
}
\email{luwm@zju.edu.cn}

\renewcommand{\shortauthors}{}

\begin{abstract}
Concomitant administration of drugs can cause drug-drug interactions (DDIs). Some drug combinations are beneficial, but other ones may cause negative effects which are previously unrecorded. Previous works on DDI prediction usually rely on hand-engineered domain knowledge, which is laborious to obtain. In this work, we propose a novel model, Molecular Substructure-Aware Network (MSAN), to effectively predict potential DDIs from molecular structures of drug pairs. We adopt a Transformer-like substructure extraction module to acquire a fixed number of representative vectors that are associated with various substructure patterns of the drug molecule. Then, interaction strength between the two drugs’ substructures will be captured by a similarity-based interaction module. We also perform a substructure dropping augmentation before graph encoding to alleviate overfitting. Experimental results from a real-world dataset reveal that our proposed model achieves the state-of-the-art performance. We also show that the predictions of our model are highly interpretable through a case study.
\end{abstract}



\maketitle

\section{Introduction}
The concurrent use of multiple drugs is inevitable when treating complex diseases, since different drugs may be targeted at different aspects of a disease. When properly combined, two drugs can act positively. However, due to laborious wet labs, many potential drug-drug interactions (DDIs) have not been documented so far, which poses a great threat to susceptible patients, especially elder people. As reported, adverse DDIs account for approximately 30\% of patient adverse events and have become an important reason for drug withdrawal \cite{Tatonetti2012ANS}. In recent years, many machine-learning and deep-neural-network-based methods have been applied to the prediction of potential DDIs.

One group of DDI prediction methods heavily relies on expert knowledge like binary drug fingerprints, which is difficult to fetch and sometimes incomplete \cite{Gottlieb2012INDIAC, Zhang2016PredictingPD, Ferdousi2017ComputationalPO}. Another group of methods requires partially known DDI networks \cite{Zhang2015LabelPP, Zhang2018ManifoldRM, Yu2018PredictingAU, Shi2019DetectingDC, ijcai2018-483}. However, when the input DDI network is very sparse, or in an extreme setting, where some drugs are isolated nodes without any known connections with other drugs, these methods will hardly produce any reliable predictions. Nyamabo {\itshape et al.} \cite{bbab133} defined this setting as “inductive”, which is much more difficult than the common “transductive” setting. The third group of methods directly exploits information from the original molecular graphs. Their input data is much easier to acquire than the former two groups. The use of raw chemical structures also makes them more flexible and expressive, being able to handle the inductive setting. Our proposed method belongs to the third group.

Among those molecular-graph-based methods, one central issue is the definition of “substructure” and the modeling of substructure interactions. Mining the interactions between the substructures of the two drugs not only helps interpret the causes of the DDIs, but also improves the prediction accuracy. Deac {\itshape et al.} \cite{deac2019drug} utilized a co-attention mechanism to jointly encode the molecular graphs of drug pairs. The co-attention mechanism is performed at node level, which can be viewed as the finest grain of substructures, but it may be suboptimal to cover the graph semantics. Also, node-level co-attention introduces substantial computational cost that is quadratic in the number of nodes, which makes it unsuitable for modeling large molecules. At a higher level, a substructure can be naturally defined as a $k$-hop neighborhood of a center node, but this definition to some extent fixes the size of the substructure and therefore is not expressive enough. Nyamabo {\itshape et al.} \cite{bbab441} defined a size-adaptive molecular substructure via learnable edge weights. However, it requires a sophisticated design of the GNN architecture, lacking flexibility. Furthermore, all of these works define a substructure around a center node, where the number of the substructures will be the same as the number of the nodes, which could be unreasonably large.

In constrast, our proposed work extracts substructures via a Transformer-like framework (MSAN-SE). We associate atoms in the molecule with a fixed number of learnable pattern vectors using attention mechanism. The substructures are adaptive and flexible in their sizes. More importantly, we define a substructure based on the cluster of a set of nodes instead of the neighborhood of a center node, and thus can generate a controllable number of substructures. After extracting substructure patterns, a similarity-based interaction module (MSAN-SI) will model the inter-graph interaction and provide pairwise interaction strength of the substructures for the final prediction module. Nevertheless, simply adopting MSAN-SE and MSAN-SI for DDI prediction may cause the model to memorize some undesirable artifacts of the dataset, resulting in the problem of overfitting. Therefore, we arrange the substructure dropping augmentation (MSAN-SD) before the GNN encoder to increase the diversity of the input data and mitigate overfitting. The data augmentation is guided by the substructure information extracted by the MSAN-SE module, which forces the model to learn a robust substructure assignment. Because our proposed modules only deal with the preprocessing and postprocessing section of the pipeline, we impose no restriction on the specific architecture of the GNN encoder. Experimental results demonstrate that our proposed method achieves the best performance in a real-world dataset.

To summarize, our main contributions are:
\begin{itemize}
\item We are the first to introduce a Transformer-like module to extract molecular substructure patterns in DDI prediction task. We further model the inter-molecule substructure interaction with a simple yet effective similarity module.
\item We introduce a novel substructure-based data augmentation method to DDI prediction task, which adjusts the model's substructure assignment and prevents the model from memorizing noisy patterns.
\item Experimental results show that our model achieves the state-of-the-art performance in both transductive and inductive settings of a real-world dataset.
\end{itemize}

\section{Methodology}

\subsection{Preliminary}

A graph $G$ can be represented by $G=(X,A)$, where 
\begin{math}
  X\in\mathbb{R}^{N\times d}
\end{math}
denotes the node feature matrix, and 
\begin{math}
  A\in\mathbb{R}^{N\times N}
\end{math}
denotes the adjacency matrix. In molecular graphs, atoms and chemical bonds are represented as nodes and edges, respectively. Graph Neural Networks (GNNs) predominantly adopt the Message Passing mechanism \cite{pmlr-v70-gilmer17a}, which aggregates information from a center node's neighborhood and updates the center node's representation. Popular GNN variants include GCN \cite{kipf2016semi}, GAT \cite{velickovic2018graph}, GIN \cite{xu2018how}, etc.

The DDI prediction task is a binary classification task on a tuple $(d_{1},d_{2},t)$ to predict whether drug $d_{1}$ and drug $d_{2}$ will interact under the DDI type $t$.

\subsection{Model Overview}

\begin{figure}
  \centering
  \includegraphics[width=\linewidth]{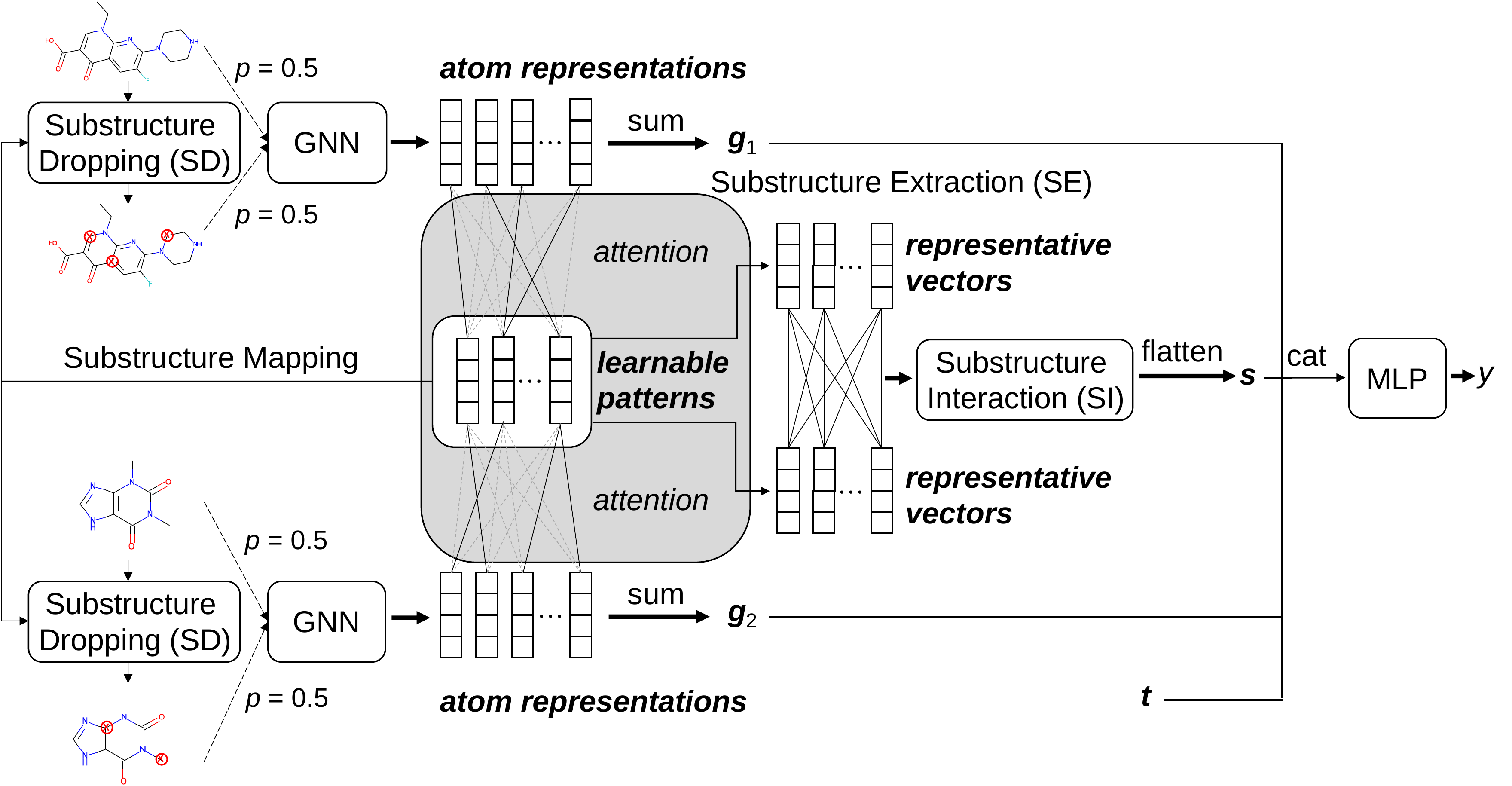}
  \caption{Framework of MSAN}
  \Description{}
  \label{fig:framework}
\end{figure}

We first randomly select half of the input molecular graphs to perform the MSAN-SD augmentation and then encode the graphs with a GNN backbone. Afterwards, an MSAN-SE module with learnable pattern vectors is employed to condense the graphs into $M$ representative vectors. Then, in the MSAN-SI module, each pair of the representative vectors from the two drugs is measured with cosine similarity, thus yielding an $M \times M$-sized similarity matrix. Finally, the flattened similarity matrix is concatenated to the representations of the two drugs and fed to an MLP layer to produce the final prediction. The framework of our model is shown in Figure~\ref{fig:framework}.

\subsection{Model Architecture}

\textbf{MSAN-SE Module.} The MSAN-SE module utilizes a Transformer-like architecture to acquire the attention scores between the queries and the keys, and then use the attention scores to fetch information from the values. It can be formulated as follows:
\begin{equation} \label{eqQKV}
  Q=Q_0W_Q,\ \ K=K_0W_K,\ \ V=V_0W_V
\end{equation}
\begin{equation} \label{eqAttnScore}
  A={\rm softmax}\left(\frac{QK^\top}{\sqrt d}\right)
\end{equation}
\begin{equation} \label{eqAttnOut}
  O={\rm ReLU}\left(\left(Q+AV\right)W_O\right)
\end{equation}
Where $Q$, $K$, $V$ denote the queries, the keys, and the values, respectively. $d$ denotes the embedding dimension. $\rm ReLU(\cdot)$ denotes ReLU activation function. $W_Q$, $W_K$, $W_V$, and $W_O$ are learnable weights. The $M$ learnable queries (patterns) in $Q_0\in\mathbb{R}^{M\times d}$ are randomly initialized, and we take the GNN-encoded node representations as our keys and values.
\begin{equation} \label{eqK0V0}
  K_0=V_0=\left[\mathbf{h}_1^{\left(L\right)},\mathbf{h}_2^{\left(L\right)},\ldots,\mathbf{h}_N^{\left(L\right)}\right]
\end{equation}
Where $L$ is the number of the GNN layers, $\mathbf{h}_i^{\left(L\right)}$ is the node $i$'s representation at the $L$-th layer, and $N$ is the number of the nodes.

Finally, we get $M$ representative vectors for each drug from Eq.(\ref{eqAttnOut}), which are associated with $M$ substructure patterns.
\begin{equation} \label{eqXrep}
  \left[\mathbf{x}_1,\mathbf{x}_2,\ldots,\mathbf{x}_M\right]=O\in\mathbb{R}^{M\times d}
\end{equation}

\textbf{MSAN-SI Module.} After we obtain the representative vectors, each pair of the representative vectors from the two drugs is measured with cosine similarity, thus yielding a similarity matrix
\begin{math}
  S\in\mathbb{R}^{M\times M}
\end{math}
\begin{equation} \label{eqCosSim}
  S_{ij}=\frac{\left(\mathbf{x}_i^{1}\right)^\top\mathbf{x}_j^{2}}{{\Vert \mathbf{x}_i^{1} \Vert_2 \Vert \mathbf{x}_j^{2} \Vert_2}}\in[-1,1]
\end{equation}
Where $\mathbf{x}_i^{1}$ is the $i$-th representative vector of drug 1, and $\mathbf{x}_j^{2}$ is the $j$-th representative vector of drug 2.

The similarity module models the interactions between the substructures of the two drugs in a non-parametric way. The values of the elements in $S$ can indicate the strength of interaction, thus making the prediction results more interpretable.

\textbf{Prediction Module.} We obtain the graph-level representation of the two drugs via a readout function. In our experiment, we use the summation readout operation
\begin{math}
  \mathbf{g}_i=\sum_{v=1}^{N}\mathbf{h}_v^i,i=1,2
\end{math}.
Where $\mathbf{h}_v^i$ denotes the node $v$’s representation of drug $i$ from the last ($L$-th) layer of the GNN.

Finally, we flatten the similarity matrix $S$, concatenate it to the representations of the two drugs, and feed them to an MLP prediction layer.
\begin{equation} \label{eqFlatten}
  \mathbf{s}={\rm flatten}(S)
\end{equation}
\begin{equation} \label{eqPredOut}
  y={\rm MLP}([\mathbf{g}_1\parallel\mathbf{g}_2\parallel\mathbf{s}\parallel\mathbf{t}])
\end{equation}
Where $\parallel$ denotes concatenation, and $\mathbf{t}$ is the one-hot encoding of a DDI type, {\itshape e.g.}, {\itshape decreased absorption}, {\itshape increased analgesic activities}, {\itshape etc.}

We adopt Binary Cross Entropy (BCE) loss as the loss function.
\begin{equation} \label{eqLoss}
  L=-\frac{1}{n}\sum_{i=1}^{n}{\left(y_i^\prime\log{\sigma(y_i)}+\left(1-y_i^\prime\right)\log{\left(1-\sigma\left(y_i\right)\right)}\right)}
\end{equation}
Where $y_i$ is the output of the $i$-th drug pair, $y_i^\prime\in\{0,1\}$ is the label of the $i$-th drug pair, $\sigma(\cdot)$ is the sigmoid function, and $n$ is the number of the drug pairs.

\textbf{MSAN-SD Module.}  Our model has 50\% probability to input an augmented version of the original graph. To augment a graph, we first compute the attention score $A$ described in Eq.(\ref{eqAttnScore}), then assign each atom to the query with the maximum probability. The substructures are defined based on the assignment. An example is shown in Figure~\ref{fig:AtomMapping}, where atoms with the same number belong to the same substructure. After that, we randomly drop a substructure from the molecule. The features of the dropped nodes are set to zero and the adjacency matrix remains unchanged.

We do not need to calculate gradients in this stage, so the computational cost is low. We do augmentation only in the training phase. The augmentation operation does not change the labels.

\section{Experiment}

\subsection{Setup}

\textbf{Datasets.} We use DrugBank \cite{bbab441} dataset in our experiment. The dataset contains 1,706 drugs and 191,808 drug pairs with 86 DDI types. Since the dataset only includes positive drug pairs, we sample the same number of negative drug pairs following \cite{bbab441}.

\textbf{Baselines.} We choose three commonly used GNN variants, GCN, GAT, and GIN as baselines. We do not apply any data augmentation and directly concatenate the graph-level representations from the two drugs to make a prediction.

We also include two newly proposed DDI prediction methods, SSI-DDI \cite{bbab133} and GMPNN \cite{bbab441} in our experiments.

\textbf{Experimental Settings.} We perform a stratified splitting to divide all the drug pairs into a training set, a validation set, and a testing set in a ratio of 6:2:2. We run the experiments on three random folds.

We also consider the inductive setting. Following \cite{bbab133}, we first split all the drugs into $\mathcal{G}_{new}$ and $\mathcal{G}_{old}$ in a ratio of 1:4. Training set $\mathcal{D}_{train}$ contains drug pairs where both of the two drugs are in $\mathcal{G}_{old}$. Testing set $\mathcal{D}_{S_1}$ contains drug pairs where both drugs are in $\mathcal{G}_{new}$. Testing set $\mathcal{D}_{S_2}$ contains drug pairs where either of the two drugs is in $\mathcal{G}_{new}$. We also repeat the experiment on three different folds.

In the inductive setting, although drugs in $\mathcal{G}_{new}$ are unseen, some common molecular substructures may exist in both $\mathcal{G}_{old}$ and $\mathcal{G}_{new}$. To make full use of drugs' substructure information, in the testing phase, the two drugs in the original drug pair are replaced with their nearest neighbors in $\mathcal{G}_{old}$ measured by Tanimoto Coefficient of ECFP \cite{ECFP}. The prediction scores of the replaced pair and the original pair are averaged to yield the final prediction score.

\textbf{Implementation details.} We run our model for 300 epochs in the training phase. We use a learning rate of 0.001 in the first 200 epochs and 0.0001 in the last 100 epochs. The batch size is set to 256. The dimension of the node embedding is tuned from \{64, 128\}. The number of representative vectors $M$ is set to 60. The initial node features include 8 categorical atomic attributes parsed from RDkit \cite{RDkit}. These attributes were then transformed into one-hot encodings and concatenated\footnote{The code is available at \url{https://github.com/Hienyriux/MSAN}.}.

\subsection{Experimental Results}

The experimental results are shown in Table~\ref{tab:DrugBankTrans}, \ref{tab:Inductive}. The experimental results demonstrate that our model achieves the state-of-the-art performance in both transductive and inductive setting.

We also report our model’s performance with different GNN backbones. We observe performance gain in almost all the backbones when applied with MSAN. Interestingly, although the GCN backbone is the simplest architecture, it outperforms the other backbones in the harder inductive setting, probably because GCN contains the minimum amount of parameters, which makes it robust to overfitting.

We observe a significant performance decline in the inductive setting. One possible reason may be the low substructure similarity between the drugs in $\mathcal{G}_{old}$ and $\mathcal{G}_{new}$. The average substructure similarity (ECFP Tanimoto Coefficient) between the drugs in $\mathcal{G}_{new}$ and their nearest neighbors in $\mathcal{G}_{old}$ is only about 0.45.

\begin{table}
  \caption{Performance (mean±std) in the transductive setting}
  \label{tab:DrugBankTrans}
  \resizebox{\linewidth}{!}{
      \begin{tabular}{lccccc}
        \toprule
        & \textbf{ACC} & \textbf{AUC} & \textbf{F1} & \textbf{P} & \textbf{R}\\
        \midrule
        SSI-DDI & 96.33±0.09 & 98.95±0.08 & 96.38±0.09 & 95.09±0.08 & 97.70±0.14\\
        GMPNN & 95.30±0.05 & 98.46±0.01 & 95.39±0.05 & 93.60±0.07 & 97.22±0.10\\
        GCN & 81.42±0.19 & 89.67±0.14 & 82.94±0.23 & 76.70±1.02 & 90.35±1.89\\
        GAT & 94.00±0.12 & 98.13±0.05 & 94.22±0.12 & 90.82±0.22 & 97.89±0.16\\
        GIN & 96.08±0.20 & 98.99±0.04 & 96.18±0.19 & 93.78±0.35 & \textbf{98.71±0.01}\\
        \hline
        MSAN-GCN (Ours) & 90.14±0.36 & 96.21±0.18 & 90.72±0.32 & 85.69±0.45 & 96.39±0.23\\
        MSAN-GAT (Ours) & 94.95±0.14 & 98.57±0.07 & 95.10±0.13 & 92.45±0.23 & 97.91±0.04\\
        MSAN-GIN (Ours) & \textbf{97.00±0.09} & \textbf{99.27±0.03} & \textbf{97.04±0.08} & \textbf{95.92±0.19} & 98.18±0.09\\
        \hline
        MSAN-GIN (w/o SE \& SI) & 95.85±0.03 & 99.02±0.03 & 95.96±0.04 & 93.33±0.02 & 98.74±0.06\\
        MSAN-GIN (w/o SD) & 96.65±0.05 & 99.13±0.03 & 96.70±0.04 & 95.52±0.08 & 97.90±0.02\\
        \bottomrule
      \end{tabular}
  }
\end{table}

\begin{table}
  \caption{Performance (mean±std) in the inductive setting}
  \label{tab:Inductive}
  \resizebox{\linewidth}{!}{
    \begin{tabular}{lcccccc}
      \toprule
      \multicolumn{1}{c}{}&
      \multicolumn{3}{c}{$\mathcal{D}_{S_1}$}&
      \multicolumn{3}{c}{$\mathcal{D}_{S_2}$}\\
      \cmidrule(r){2-4}
      \cmidrule(r){5-7}
      & \textbf{ACC} & \textbf{AUC} & \textbf{F1} & \textbf{ACC} & \textbf{AUC} & \textbf{F1}\\
      \midrule
      SSI-DDI & 65.40±1.30 & 73.43±1.81 & 54.12±3.46 & 76.38±0.92 & 84.23±1.05 & 73.54±1.50\\
      GMPNN & 68.57±0.30 & 74.96±0.40 & 65.32±0.23 & 77.72±0.30 & 84.84±0.15 & \textbf{78.29±0.16}\\
      GCN & 65.28±0.74 & 71.01±0.93 & 65.78±1.45 & 70.78±0.56 & 77.39±0.88 & 71.65±1.39\\
      GAT & 65.65±0.96 & 71.46±0.89 & 65.82±2.20 & 74.07±0.34 & 81.63±0.43 & 71.97±1.30\\
      GIN & 63.47±1.01 & 69.08±1.11 & 56.00±1.00 & 74.56±0.36 & 82.14±0.41 & 72.25±0.78\\
      \hline
      MSAN-GCN (Ours) & \textbf{69.17±1.04} & \textbf{76.12±0.89} & \textbf{67.10±2.17} & \textbf{77.81±0.48} & \textbf{85.74±0.27} & 76.48±0.94\\
      MSAN-GAT (Ours) & 67.73±1.29 & 75.51±1.66 & 62.01±2.62 & 77.42±0.72 & 85.06±0.57 & 75.97±1.22\\
      MSAN-GIN (Ours) & 65.24±1.25 & 71.96±1.74 & 60.26±1.97 & 74.98±0.56 & 84.10±0.54 & 71.06±1.46\\
      \bottomrule
    \end{tabular}
  }
\end{table}

\subsection{Ablation Study}

We conduct an ablation study to verify the effectiveness of the MSAN module. As is shown in Table~\ref{tab:DrugBankTrans}, the MSAN-SE and MSAN-SI module significantly improves the precision score by 2.59, while it drops the recall score by 0.56. The result demonstrates that the MSAN-SE and MSAN-SI module can boost the model's performance by fully exploiting the interaction patterns in the existing training data, but at the risk of impairing the model's generalization ability. Nevertheless, we can also find that both the precision score and the recall score increase with the MSAN-SD augmentation, which indicates the augmentation can alleviate overfitting.

\subsection{Case Study}

MSAN-SE module calculates the mapping probability between each atom in a molecule and each learnable query through attention mechanism. Here, we assign each atom to the query with the maximum probability. The substructures are defined based on the assignment. Also, the similarity matrix $S$ offers information about the strength of substructure interactions between the two drugs, which can explain the mechanism behind the DDI.

\begin{figure}
  \centering
  \includegraphics[width=\linewidth]{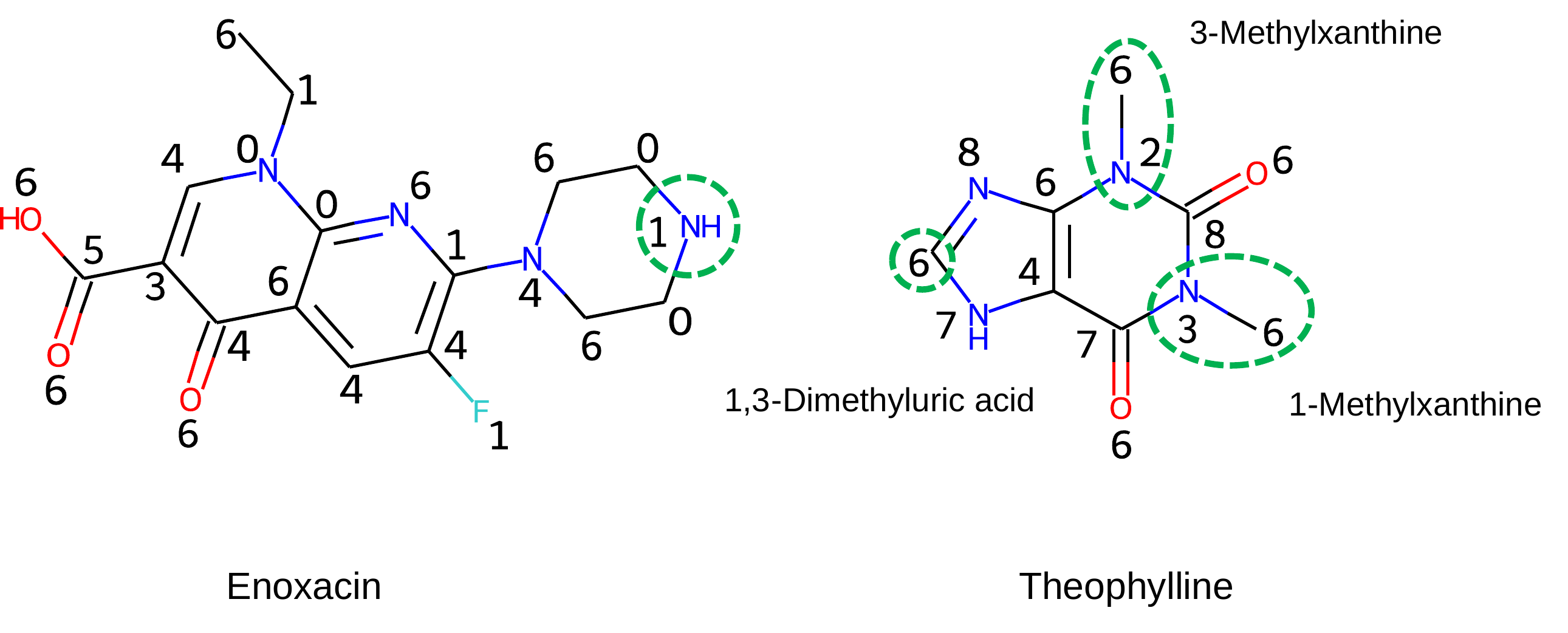}
  \caption{Atom mapping of Enoxacin and Theophylline}
  \Description{}
  \label{fig:AtomMapping}
\end{figure}

Here, we take the interaction between Enoxacin and Theophylline as an example. According to our model’s prediction, these two drugs have a probability of 0.99 to interact under the DDI type {\itshape decreased metabolism}. The atom mapping ($M=10$) of Enoxacin and Theophylline is shown in Figure~\ref{fig:AtomMapping}. The numbers close to the atoms are the indices of the learnable queries (patterns). Atoms with the same number belong to the same substructure.

Enoxacin is an oral broad-spectrum fluoroquinolone antibacterial agent. Theophylline is a phosphodiesterase inhibiting drug for respiratory diseases. Enoxacin can inhibit the cytochrome P-450-mediated Theophylline metabolism, resulting in reduced metabolic clearance of Theophylline, which further increases the incidence of cardiac and central nervous system adverse reactions \cite{MizukiEnoxacin}.

Research \cite{MizukiEnoxacin} shows that the basicity of the 4’-nitrogen atom (green dashed circled) of Enoxacin is a possible factor in the inhibition of Theophylline metabolism. On the other side, Theophylline has three typical metabolites associated with three positions in the green dashed circles. Therefore, the most relevant substructures of the DDI should be Substructure 1, 0 of Enoxacin, and Substructure 2, 3, 6 of Theophylline. In Table~\ref{tab:SubInt}, six out of ten interactions involve Enoxacin Substructure 1 or 0, the top six interactions with the highest scores all involve Theophylline Substructure 2 or 3, and one interaction contains Theophylline Substructure 6. The analysis of substructure mapping and interaction strength indicates that our model can highlight important parts of drug molecules, thus being highly explainable.

\begin{table}
  \caption{Substructure interactions with top-10 scores
between Enoxacin and Theophylline}
  \label{tab:SubInt}
  \resizebox{0.6\linewidth}{!}{
      \begin{tabular}{cccc}
        \toprule
        Rank & \begin{tabular}{c} Substructure in \\ Enoxacin \end{tabular} & \begin{tabular}{c} Substructure in \\ Theophylline \end{tabular} & Score\\
        \midrule
        1 & 1 & 2 & 0.619\\
        2 & 3 & 2 & 0.579\\
        3 & 0 & 2 & 0.529\\
        4 & 1 & 3 & 0.521\\
        5 & 3 & 3 & 0.518\\
        6 & 0 & 3 & 0.443\\
        7 & 1 & 4 & 0.438\\
        8 & 6 & 6 & 0.399\\
        9 & 0 & 4 & 0.365\\
        10 & 3 & 4 & 0.331\\
        \bottomrule
      \end{tabular}
  }
\end{table}

\subsection{Potential DDI prediction}

The DDI dataset is incomplete and often fails to include many potential DDIs. To demonstrate our model’s capability of predicting potential DDIs, in Table~\ref{tab:Potential}, we provide ten verified positive DDIs which do not appear in DrugBank dataset or are wrongly labeled as negative in the negative sampling process. The evidence of each potential DDI is retrieved from DrugBank website (\url{https://go.drugbank.com}) which includes a considerable number of newly documented DDIs that are not found in the original DrugBank dataset.

\begin{table}
  \caption{Potential DDIs found by our model}
  \label{tab:Potential}
  \resizebox{\linewidth}{!}{
    \begin{tabular}{lllc}
      \toprule
      Drug 1 & Drug 2 & Description & Probability\\
      \midrule
      Tianeptine & Levomilnacipran & increased risk or severity of adverse effects & 0.9996\\
      Norelgestromin & Glimepiride & decreased therapeutic efficacy & 0.9996\\
      Simvastatin & Valsartan & decreased metabolism & 0.9976\\
      Milnacipran & Indacaterol & increased tachycardic activities & 0.9969\\
      Molsidomine & Amyl Nitrite & increased hypotensive activities & 0.9955\\
      Alimemazine & Ephedrine & increased therapeutic efficacy & 0.9949\\
      Fenofibrate & Ethanol & increased myopathic rhabdomyolysis activities & 0.9896\\
      Tacrolimus & Gemifloxacin & increased risk or severity of QTc prolongation & 0.8831\\
      Diltiazem & Pinaverium & increased arrhythmogenic activities & 0.8416\\
      Methohexital & Mifepristone & increased metabolism & 0.7275\\
      \bottomrule
    \end{tabular}
  }
\end{table}

\section{Conclusion}

In this paper, we propose a new framework MSAN to predict drug-drug interactions based on drugs’ molecular structures. We adopt a Transformer-like architecture to obtain a drug’s substructures and employ a similarity module to model the interaction strength between the two drugs’ substructures. To increase the diversity of the training data and promote robust extraction of substructures, we also introduce a substructure dropping augmentation. Empirical results from the real-world dataset illustrate the advanced performance of our model. Our model is also highly interpretable and able to discover potential DDIs. Our framework is possible to be further applied to other graph-interaction-related tasks.

\begin{acks}
This work is supported by the National Key Research and Development Project of China (No. 2018AAA0101900), the Key Research and Development Program of Zhejiang Province, China (No. 2021C01013), CKCEST, and MOE Engineering Research Center of Digital Library.
\end{acks}

\clearpage

\balance

\bibliographystyle{ACM-Reference-Format}
\bibliography{sample-base}

\appendix

\end{document}